\documentclass{IOS-Book-Article}
\usepackage{float}
\usepackage{mathptmx}
\usepackage{soul}\setuldepth{article}

\def\hb{\hbox to 11.5 cm{}}
\usepackage{times}
\usepackage{latexsym}

\usepackage{float}
\usepackage{booktabs}
\usepackage{amsmath}
\usepackage{enumitem}
\usepackage{multirow}
\usepackage{graphicx}
\usepackage{subcaption}
\usepackage{placeins}
\usepackage{fvextra}
\usepackage[utf8]{inputenc}
\UseRawInputEncoding
\usepackage{array}
\newcolumntype{L}[1]{>{\raggedright\arraybackslash}p{#1}}

\begin{document}

\pagestyle{headings}
\def\thepage{}
\begin{frontmatter}              

\title{It Depends on the Task: Rethinking Rhetorical Segmentation of Dutch Judicial Decisions}

\author[A]{\fnms{Samyak} \snm{Sheth}\orcid{0009-0005-8633-825X}%
\thanks{Corresponding Author: Samyak Sheth, samyak.sheth@maastrichtuniversity.nl}},
\author[A]{\fnms{Gerasimos} \snm{Spanakis}\orcid{0000-0002-0799-0241}}
and
\author[A]{\fnms{Gijs} \snm{van Dijck}\orcid{0000-0003-4102-4415}}

\address[A]{Maastricht University, the Netherlands.}

\begin{abstract}
Classifying the material facts, submissions, judicial reasoning, and the court's decision (rhetorical roles) in case law is a prerequisite for high-quality performance across virtually all NLP-in-law tasks, including legal information retrieval, summarisation, and LLM-based analysis. These rhetorical roles often intertwine and shift sentence by sentence, making reliable classification difficult. This paper introduces the first sentence-level annotated dataset of Dutch judicial decisions, labelled across four legally meaningful categories: material facts, submissions, judicial reasoning, and the court's decision, alongside non-annotated (None) content, and compares several classification strategies, including a staged pipeline that separates None detection from the rhetorical roles, and a Bayesian update that leverages information about the section-header a sentence belongs to. Our results show that no single strategy dominates on both precision and recall: header priors help most when the None sentences are excluded and offer little benefit for None detection itself. These trade-offs matter in practice. For instance, extractive summarisation depends more on recall so it does not miss relevant facts, whereas outcome prediction depends more on precision regarding decision and facts sentences so that the predicted ruling is not contaminated by noise. We use these trade-offs and map the approaches tested in this study onto a set of representative legal NLP use cases.

\end{abstract}

\begin{keyword}
Legal NLP, rhetorical role classification, judicial decisions, Dutch NLP, sentence classification, Bayesian updating, information extraction, legal information retrieval
\end{keyword}
\end{frontmatter}

\section{Introduction}

\begin{figure*}[h]
    \centering     
    \includegraphics[width=0.8\textwidth]{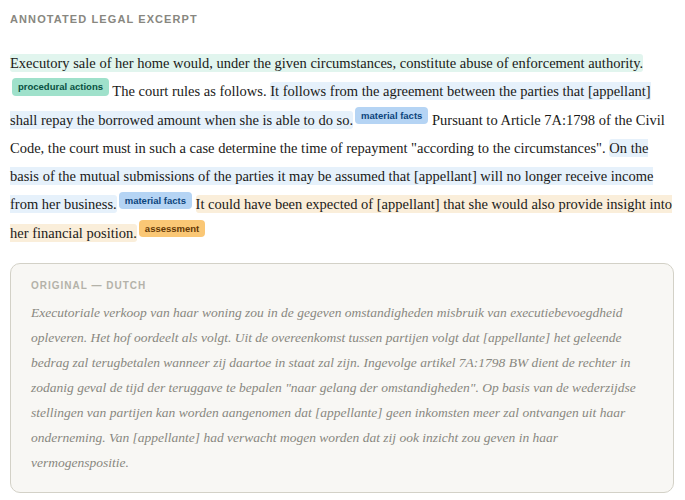}
    \caption{An annotated excerpt from the assessment section (beoordeling) of a Dutch judicial decision, illustrating how material facts, procedural actions, and court assessments interleave within a single passage along with unlabelled (not related) content.}
    \label{fig:annotation}
\end{figure*}

Nearly every computational task in the legal domain depends on knowing what role a sentence plays in a judicial decision  \cite{walker2019automatic, bhattacharya2019identification, malik-etal-2022-semantic, nigam-etal-2025-legalseg}. Retrieval systems that search specifically within a factual background \cite{lou2021extracting, shulayeva2017recognizing} or reasoning section \cite{csanyi2025sentence} return more targeted results than those treating decisions as undifferentiated text. Summaries that distinguish facts from reasoning and conclusions outperform those that do not \cite{zhong2022computing, kalamkar-etal-2022-corpus}. For LLM-based systems \cite{magesh2025hallucination, dahl2024large}, whether fine-tuned or prompted \cite{chalkidis2020legalbert, deroy2024applicability}, absent segmentation means the model cannot reliably distinguish a court's judgment from its reasoning or procedural context--a bottleneck restricting everything built downstream \cite{ariai2024survey, kalamkar-etal-2022-corpus}.

High scores on benchmark corpora suggest that the classification of rhetorical roles in judicial decisions is largely tractable \cite{walker2019automatic, bhattacharya2019identification, savelka2018segmenting, malik-etal-2022-semantic, kalamkar-etal-2022-corpus, nigam-etal-2025-legalseg, santosh2023joint, santosh2024mind, csanyi2025sentence, darji2025segmentation, lou2021extracting, xu2021}. In practice, it is not. In our corpus, document sections are frequently not homogeneous in rhetorical role, and a sentence often has a different role from the sentence preceding it. Figure~\ref{fig:annotation} illustrates that the categories of interest do not cluster neatly within document sections, and that the facts are often intertwined with what parties claim or submit and what the court reasons and decides. As a result, a sentence can often not be labelled from its position alone. Dutch case law has no annotated corpus of this kind.

Moreover, what to prioritize when segmenting depends on the task. For instance, for summarization of the facts one may prioritize not missing relevant facts over the inclusion of some noise (e.g., non-facts). Yet for retrieving relevant legislative provisions based on the facts it can be more important to avoid noise and prioritize high precision levels.

In this study, we explore approaches that work, even if they are not the newest. We apply the approaches to do sentence-level extraction of legally relevant information from Dutch judicial decisions across four core categories: \textit{material facts} (\textit{materi\"{e}le feiten}), \textit{submissions} (\textit{proceshandelingen}), \textit{judicial reasoning} (\textit{beoordeling}), and \textit{court's decision} (\textit{beslissing}), alongside a residual class of non-annotated content. 

We find that no single strategy is best on both precision and recall, so the choice depends on the downstream task: based on these profiles, recall-oriented tasks such as summarisation and precision-oriented tasks such as provision retrieval would favour different formulations. We make three contributions: (1) the first sentence-level annotated dataset of Dutch judicial decisions labelled for rhetorical role, with inter-annotator agreement of $\kappa$ = 0.799–-0.962 (mean 0.871) and released publicly; (2) a comparison of classification strategies, including a direct five-way classifier, staged pipelines that separate None detection from role assignment, and header-based Bayesian updating; (3) an assessment of which strategy suits which downstream aim.

\section{Related Work}
Malik et al.~\cite{malik-etal-2022-semantic} established rhetorical role classification as a well-defined NLP task with a large-scale annotated corpus of Indian legal judgments across 13 roles, serving as a reference point for subsequent work. Kalamkar et al.~\cite{kalamkar-etal-2022-corpus} extended this with a corpus targeting summarisation and judgment prediction. Nigam et al.~\cite{nigam-etal-2025-legalseg} scaled these efforts with LegalSeg (7,000+ documents, 1.4M sentences), finding that models incorporating broader context consistently outperform sentence-level-only approaches. At SemEval-2023, Tandon and Chatterjee~\cite{tandon-chatterjee-2023-lrl-nc} combined topic modelling with RoBERTa and a CRF layer, achieving micro-F1 of 0.798.
A recurring finding is that classification improves when structural cues are incorporated. Xu et al.~\cite{xu2021} showed that sentence position combined with domain-specific embeddings improves classification into issues, conclusions, and reasons---motivating our Bayesian structural update, which treats section headers as a probabilistic signal rather than a hard override. Santosh et al.~\cite{santosh2023joint} reformulated the task at span level using a semi-Markov CRF, improving over sentence-level baselines, with a span-to-sentence mapping parallel to our annotation design. In later work, Santosh et al.~\cite{santosh2024mind} showed that semantically similar instances at inference time yield consistent macro-F1 gains. Walker et al.~\cite{walker2019automatic} found that rule-based approaches plateau on larger datasets where learned models continue to improve, informing our fully data-driven pipeline.
The work most directly comparable to ours is that of Cs\'{a}nyi et al.~\cite{csanyi2025sentence}, who studied sentence-level rhetorical role labelling for Hungarian judicial decisions with iterative human annotation, evaluating a range of neural architectures and embedding strategies. Our work differs in language (Dutch), in explicitly distinguishing None Detection (i.e., first filtering sentences that should receive a (any) label) from role classification, and in introducing a Bayesian structural update component. Lou et al.~\cite{lou2021extracting} offer a complementary perspective on French decisions, applying paragraph segmentation to extract case-relevant facts and distinguishing what a sentence says from the function it serves: a distinction central to our annotation scheme. For Dutch legal text specifically, van Drie et al.~\cite{van2023dutch} introduced a sentence-level annotated dataset of legislative text labelled for semantic roles; though focused on statutory rather than judicial text, the annotation challenges they identify stylistic variability, domain vocabulary, and sparse labelling---closely reflect those encountered in our study.

\section{Dataset}
\subsection{Corpus Scope \& Annotation}

The experimental corpus initially contained 102 Dutch judicial decisions sampled from Rechtspraak.nl, the official public portal of the Dutch judiciary. The decisions cover the 2015-2021 period and three major areas of Dutch law: civil, criminal, and administrative law. Cases were randomly sampled with approximately equal representation across the three domains and 16 distinct court types. Following annotation, two decisions were excluded because their original XML files could not be retrieved. The final dataset hence contains 100 decisions.

The annotation scheme contains four rhetorical roles: \textit{materi\"ele feiten} (the substantive factual basis of the dispute); \textit{proceshandelingen} (parties' positions and formal procedural acts); \textit{beoordeling} (the court's legal reasoning); and \textit{beslissing} (the court's final ruling). Sentences that did not belong to one of these four roles were assigned \textit{None}. The four roles mirror the section types that conventionally structure Dutch judicial decisions. We did not adopt finer-grained schemes, as several fine-grained classes would become too sparse to learn, and beslissing alone already covers only 2.9\% of sentences as reported in Table \ref{tab:label_distribution}.

One law professor and one PhD student in law annotated the dataset, with each being assigned 51 decisions. Double-annotating the full corpus would have doubled the workload of two legal experts, so agreement was instead assessed on a separate set of 20 decisions, roughly one fifth of the corpus size. After eleven rounds of annotation, discussion, and guideline refinement, both annotators independently annotated these 20 decisions, which are not included in the experimental corpus. Agreement was evaluated using length-weighted word-level Cohen's $\kappa$ and span-level F1, where a span was a match when one annotator's span was fully contained within the other's. Word-level agreement was high across the four labels ($\kappa=0.799$--$0.962$; mean $=0.871$), and mean span-level F1 0.790 under contained matching. Full per-label agreement results are provided in the accompanying \href{https://github.com/SamyakSheth/Rhetorical-Role-Extraction}{repository}. In total, annotation required an estimated 80--115 person-hours of expert time, including at least 44 hours across the eleven guideline-refinement rounds.

\subsection{Preprocessing and Dataset Statistics}

\begin{table}[h]
\centering
\caption{Label distribution in the annotated dataset.}
\label{tab:label_distribution}
\begin{tabular}{lrr}
\hline
\textbf{Label} & \textbf{Count} & \textbf{Proportion} \\
\hline
None & 3{,}060 & 36.5\% \\
beoordeling & 2{,}293 & 27.3\% \\
materi\"{e}le feiten & 1{,}529 & 18.2\% \\
proceshandelingen & 1{,}263 & 15.1\% \\
beslissing & 246 & 2.9\% \\
\hline
Total & 8{,}391 & 100.0\% \\
\hline
\end{tabular}
\end{table}

The XML documents were converted to plain text while preserving section boundaries and their corresponding headers. A custom rule-based splitter for Dutch legal text was used that accounted for common legal abbreviations, enumerations, and irregular formatting. Raw section headers were subsequently normalized into five functional groups: \textit{Context}, \textit{Feiten}, \textit{Beoordeling}, \textit{Beslissing}, and \textit{Proceshandelingen partijen}. Sections corresponding to \textit{Wettelijk kader}, \textit{Proceskosten}, and \textit{Procesverloop} were removed from the corpus because they fell outside the scope of the annotation task. Full preprocessing rules and header mappings are provided with the accompanying code.

During preprocessing, 70 malformed sentence units (0.83\% of 8,461) could not be reliably associated with valid sentence boundaries and structural metadata and were therefore excluded, leaving 8,391 sentences in the final dataset. The data were split at the case level to prevent information leakage, resulting in 5,608 training, 1,281 validation, and 1,502 test sentences. Table~\ref{tab:label_distribution} reports the final label distribution.

The dataset constitutes a core contribution of this study and is publicly available at \url{https://github.com/SamyakSheth/Rhetorical-Role-Extraction}. It contains the sentence-level annotations, the ECLI identifiers of all 100 decisions, the structural metadata, annotation guidelines, and the complete code with additional results. The original decisions can be retrieved in XML format from the Open Data service of the Dutch judiciary \url{https://www.rechtspraak.nl/uitspraken/open-data} using the provided ECLI identifiers.

\section{Methodology}

\begin{figure*}[h]
\centering
\includegraphics[width=0.90\textwidth]{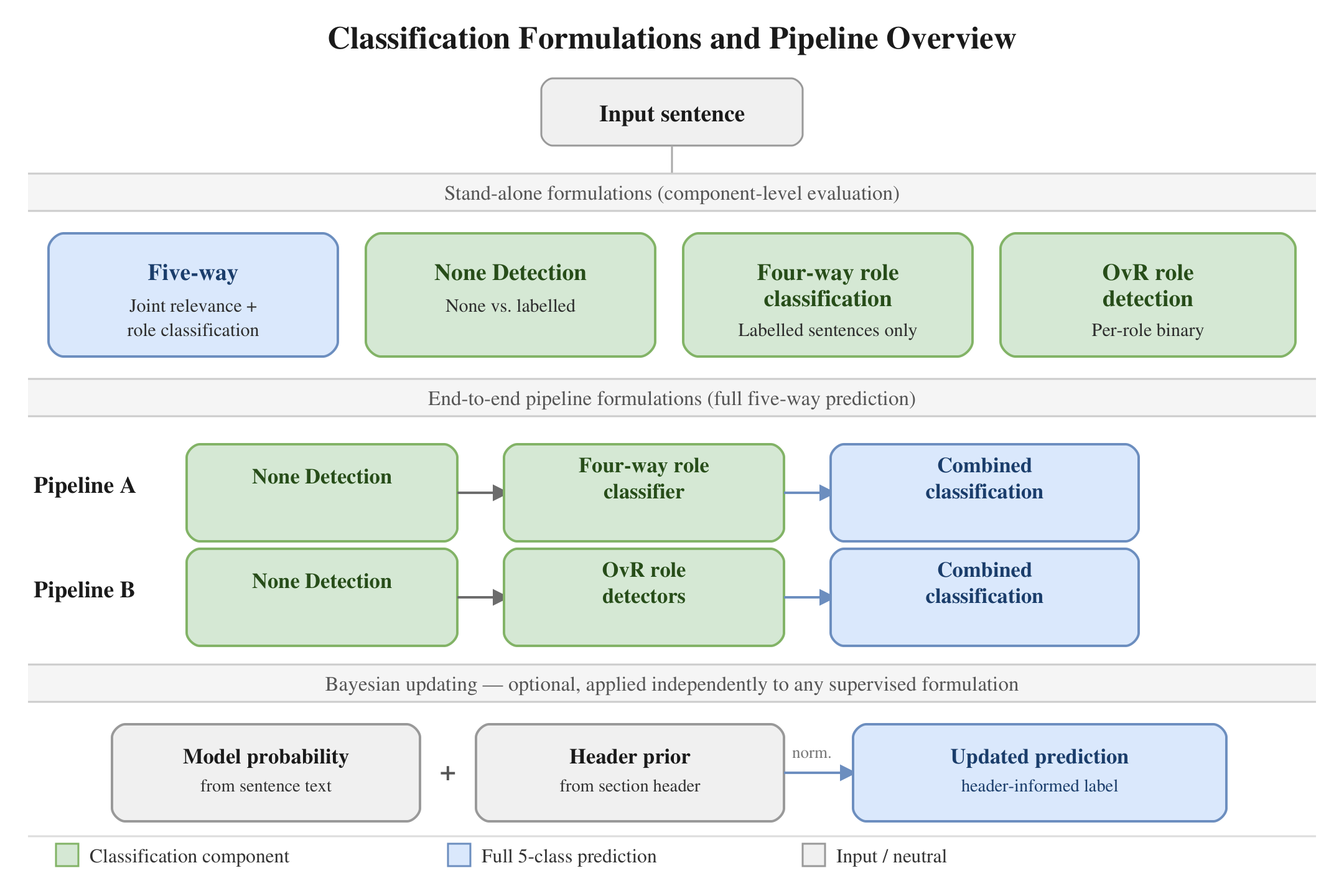}
\caption{Overview of classification formulations, end-to-end pipeline combinations, and Bayesian updating. Green boxes indicate classification components; blue boxes indicate final five-way predictions; grey boxes indicate inputs.}
\label{fig:pipeline_overview}
\end{figure*}

\subsection{Classification Task Setup}

The annotation scheme defines a five-label task: four substantive legal-role categories plus \texttt{None}. We formulate the problem in 2 ways: (a) \textbf{Five-way Classification} predicts one of the five labels directly and (b) a pipeline approach: \textbf{None Detection} is a binary classifier separating \texttt{None} from the four substantive categories. Next, a \textbf{Four-Way Role Classification} excludes \texttt{None} sentences entirely, training and evaluating only on gold legally-labelled sentences. Alternatively, \textbf{One-vs-Rest Role Detection (OvR)} trains one independent binary classifier per legal role; unlike the other formulations, this allows a sentence to receive multiple labels.


Three model families were evaluated. Logistic Regression with TF-IDF features~\cite{pedregosa2011scikit} served as a lexical baseline. Two Dutch encoder-based language models, RobBERT~\cite{delobelle-etal-2020-robbert} and BERTje~\cite{devries2019bertje}, were fine-tuned separately for each formulation on the same frozen train/validation/test splits. Three open-source instruction-tuned LLMs — Gemma-3-12B-IT~\cite{gemmateam2025gemma3}, Reynaerde-7B-Chat~\cite{vandenavenne2024reynaerde}, and Qwen2.5-7B-Instruct~\cite{qwenteam2024qwen25} — were evaluated zero-shot, few-shot, and with header-aware few-shot prompting, without fine-tuning. Prompt wording was iterated but did not materially affect performance; final templates and results are in the \href{https://github.com/SamyakSheth/Rhetorical-Role-Extraction}{GitHub repository}. No legal-domain model was included, as, to our knowledge, most available legal language models are pretrained predominantly on English text.

\subsection{End-to-End Pipeline Formulations}

Figure~\ref{fig:pipeline_overview} shows how the component formulations combine into two end-to-end pipelines that preserve Stage~1's uncertainty on None Detection. Pipeline A combines None Detection with Four-Way Role Classification: with $z \in \{0,1\}$ indicating relevance, $P(\texttt{None} \mid s) = P(z{=}0 \mid s)$ and $P(y \mid s) = P(z{=}1 \mid s)\,P(y \mid s, z{=}1)$ for each legal role $y$. Pipeline B instead uses the OvR detectors: since these are trained independently, their positive-class scores are normalised across the four labels before being weighted by $P(z{=}1 \mid s)$.

\subsection{Header-Based Bayesian Updating}

For each header group $h$ we estimate from the training data how often each label occurs under that header, yielding a prior $P(y \mid h)$---a lightweight, parameter-free way of incorporating document structure that also isolates its contribution from the underlying classifier. At inference, the prior is combined with the text-model probability as $\log P(y \mid s,h) \propto \log P(y \mid s) + \lambda \log P(y \mid h)$; scores are renormalised and the highest-scoring label is selected. $\lambda$ is tuned separately for each model and formulation on the validation set, sweeping $\lambda \in \{0, 0.1, 0.25, 0.5, 1, 2, 3, 5\}$ and selecting the value that maximises macro F1. Pipeline components are tuned independently, and each of Pipeline B's four OvR detectors receives its own $\lambda$. Bayesian updating applies only to models with probabilistic outputs; tuned values for $\lambda$ are reported in Table~\ref{app:full_supervised_results}.

\section{Experiments \& Results}

\subsection{Overall Results}

Table~\ref{app:full_supervised_results} reports the results for each supervised model before and after header-prior fusion, with macro F1 as the main comparison metric given the imbalanced label distribution. The Dutch encoder models outperform both the Logistic Regression baseline and prompting-based LLMs throughout (full results for both in the accompanying \href{https://github.com/SamyakSheth/Rhetorical-Role-Extraction}{repository}): RobBERT reaches 0.6680 macro F1 before fusion in the direct five-way setting against 0.6561 for BERTje, and contextual representations outperform the lexical and prompting baselines across every other setting too.
    \begin{table*}[!t]
    \centering
    \caption{Overall supervised model performance before and after Bayesian updating}
    \label{app:full_supervised_results}
    \scriptsize
    \setlength{\tabcolsep}{2.5pt}
    \renewcommand{\arraystretch}{1}
    \resizebox{\textwidth}{!}{%
    \begin{tabular}{lllrccccc|ccccc}
    \toprule
    \multirow{2}{*}{\textbf{Model}} 
    & \multirow{2}{*}{\textbf{Setting}} 
    & \multirow{2}{*}{\textbf{Subset}} 
    & \multirow{2}{*}{\textbf{N}} 
    & \multicolumn{5}{c|}{\textbf{Before Bayes}} 
    & \multicolumn{5}{c}{\textbf{After Bayes}} \\
    \cmidrule(lr){5-9} \cmidrule(lr){10-14}
    & & & 
    & \textbf{Acc.} & \textbf{P} & \textbf{R} & \textbf{F1} & \textbf{W-F1}
    & \textbf{Acc.} & \textbf{P} & \textbf{R} & \textbf{F1} & \textbf{W-F1} \\
    \midrule
    \multirow{7}{*}{BERTje}
    & 5-way & incl.\ None & 1502 & 0.6458 & 0.6774 & 0.6398 & 0.6561 & 0.6442 & 0.6551 & 0.6890 & 0.6416 & 0.6615 & 0.6528 \\
    & Pipeline A & incl.\ None & 1502 & 0.6258 & 0.6466 & 0.5997 & 0.6164 & 0.6235 & 0.6505 & 0.6835 & 0.6300 & 0.6496 & 0.6487 \\
    & Pipeline B & incl.\ None & 1502 & 0.6272 & 0.6530 & 0.6028 & 0.6212 & 0.6256 & 0.6585 & 0.6900 & 0.6474 & 0.6625 & 0.6573 \\
    & Stage 1 & incl.\ None (binary) & 1502 & 0.7783 & 0.7595 & 0.7589 & \textbf{0.7592} & 0.7782 & 0.7783 & 0.7596 & 0.7581 & 0.7588 & 0.7780 \\
    & Stage 2 4-way & excl.\ None & 961 & 0.7138 & 0.7484 & 0.7128 & 0.7272 & 0.7130 & 0.7638 & 0.7996 & 0.7619 & \textbf{0.7781} & 0.7626 \\
    & OvR mean & excl.\ None & 961 & 0.8606 & 0.8336 & 0.7965 & 0.8118 & 0.8579 & 0.8824 & 0.8767 & 0.8050 & 0.8298 & 0.8761 \\
    & OvR mean & incl.\ None & 1502 & 0.8399 & 0.7514 & 0.7708 & 0.7531 & 0.8464 & 0.8694 & 0.7807 & 0.7786 & \textbf{0.7772} & 0.8704 \\
    \midrule
    \multirow{7}{*}{RobBERT}
    & 5-way & incl.\ None & 1502 & 0.6658 & 0.6911 & 0.6515 & 0.6680 & 0.6634 & 0.6758 & 0.7069 & 0.6566 & \textbf{0.6772} & 0.6732 \\
    & Pipeline A & incl.\ None & 1502 & 0.6372 & 0.6615 & 0.6498 & 0.6513 & 0.6380 & 0.6458 & 0.6714 & 0.6593 & \textbf{0.6609} & 0.6461 \\
    & Pipeline B & incl.\ None & 1502 & 0.6372 & 0.6705 & 0.6567 & 0.6604 & 0.6377 & 0.6551 & 0.6879 & 0.6820 & \textbf{0.6810} & 0.6565 \\
    & Stage 1 & incl.\ None (binary) & 1502 & 0.7856 & 0.7750 & 0.7468 & 0.7562 & 0.7799 & 0.7856 & 0.7750 & 0.7468 & 0.7562 & 0.7799 \\
    & Stage 2 4-way & excl.\ None & 961 & 0.7315 & 0.7697 & 0.7410 & 0.7526 & 0.7304 & 0.7492 & 0.7889 & 0.7546 & 0.7687 & 0.7481 \\
    & OvR mean & excl.\ None & 961 & 0.8733 & 0.8472 & 0.8224 & 0.8328 & 0.8709 & 0.8832 & 0.8674 & 0.8236 & \textbf{0.8394} & 0.8786 \\
    & OvR mean & incl.\ None & 1502 & 0.8479 & 0.7583 & 0.7957 & 0.7683 & 0.8542 & 0.8620 & 0.7681 & 0.7957 & \textbf{0.7772} & 0.8656 \\
    \bottomrule
    \end{tabular}%
    }
    \begin{flushleft}
    \footnotesize
    \textit{Note.} P, R and F1 are macro-averaged; W-F1 is weighted F1; N is the number of test sentences. Stage 1 is binary (None vs.\ labelled); Stage 2 is evaluated on gold-labelled sentences only. OvR mean averages the four detectors' both-class (target vs.\ not-target) scores; ``incl.\ None'' scores the detectors on all test sentences, counting every non-target sentence as a negative. ``After Bayes'' uses $\lambda$ tuned per component on the validation set, listed as 5-way/Stage~1/Stage~2/OvR (OvR in the order \textit{beoordeling}, \textit{beslissing}, \textit{materi\"ele feiten}, \textit{proceshandelingen}): RobBERT $0.5/0/0.5/(2.0, 0.0, 2.0, 0.1)$; BERTje $0.25/0.5/1.0/(5.0, 0.0, 3.0, 1.0)$. Pipelines A/B reuse their components' values.
    \end{flushleft}
\end{table*}

Header-prior fusion has a task-dependent effect (Table~\ref{app:full_supervised_results}). It modestly helps direct five-way classification for both encoders, does not help None Detection (the sweep selects $\lambda$ = 0 for RobBERT, and BERTje $\lambda$ = 0.5 yields no test gain), and substantially improves four-way role classification and, more modestly, OvR role detection. The pipelines confirm this within a single model: RobBERT fused Pipeline B (0.6810) exceeds both the unfused (0.6680) and fused (0.6772) five-way results, and BERTje pipelines gain even more under fusion.Scored on all test sentences, including None, fused OvR macro F1 drops to 0.777 for both encoders (from 0.839 for RobBERT and 0.830 for BERTje), showing the cost of applying the detectors without a None filter.

We also tested a confidence-thresholding decision rule: predict \textit{None} only once $P(\text{None})$ clears a threshold $\theta$, otherwise defer to the role model (and, for Pipeline B, to per-detector thresholds on the OvR outputs), with $\theta$ tuned on the validation set to maximize each class's own binary F1 in isolation. This underperformed header-prior fusion in every formulation tested, and it also fell below the unfused baseline for the five-way model and both pipelines (five-way: 0.6680 unfused, 0.6772 fused, 0.6240 thresholded; Pipeline A: 0.6513 unfused vs.\ 0.5983 thresholded; Pipeline B: 0.6604 unfused vs.\ 0.6056 thresholded).

\subsection{Class-Level Analysis}

\begin{table*}[!t]
\centering
\caption{RobBERT class-level results across formulations, before and after header-prior fusion. OvR rows report each detector's own positive-class precision/recall/F1. $\lambda$ values as in Table 2.}
\label{tab:classwise_before_after}
\scriptsize
\setlength{\tabcolsep}{3pt}
\renewcommand{\arraystretch}{0.95}
\resizebox{\textwidth}{!}{%
\begin{tabular}{llrcccccc}
\toprule
\multirow{2}{*}{\textbf{Formulation}} & \multirow{2}{*}{\textbf{Class}} & \multirow{2}{*}{\textbf{Support}}
& \multicolumn{3}{c}{\textbf{Before (plain)}} & \multicolumn{3}{c}{\textbf{After (tuned $\lambda$)}} \\
\cmidrule(lr){4-6} \cmidrule(lr){7-9}
& & & \textbf{P} & \textbf{R} & \textbf{F1} & \textbf{P} & \textbf{R} & \textbf{F1} \\
\midrule

\multirow{5}{*}{5-way}
& None & 541 & 0.7033 & 0.7579 & 0.7295 & 0.6933 & 0.7689 & 0.7292 \\
& beoordeling & 389 & 0.6355 & 0.6992 & 0.6659 & 0.6548 & 0.7069 & 0.6799 \\
& beslissing & 69 & 0.8947 & 0.7391 & 0.8095 & 0.9074 & 0.7101 & 0.7967 \\
& materi\"ele feiten & 297 & 0.6423 & 0.5320 & 0.5820 & 0.6639 & 0.5387 & 0.5948 \\
& proceshandelingen & 206 & 0.5798 & 0.5291 & 0.5533 & 0.6150 & 0.5583 & 0.5852 \\
\midrule

\multirow{5}{*}{Pipeline A}
& None & 541 & 0.7403 & 0.6322 & 0.6820 & 0.7370 & 0.6266 & 0.6773 \\
& beoordeling & 389 & 0.5706 & 0.7378 & 0.6435 & 0.5787 & 0.7558 & 0.6555 \\
& beslissing & 69 & 0.8438 & 0.7826 & 0.8120 & 0.8438 & 0.7826 & 0.8120 \\
& materi\"ele feiten & 297 & 0.6331 & 0.5286 & 0.5761 & 0.6494 & 0.5488 & 0.5949 \\
& proceshandelingen & 206 & 0.5200 & 0.5680 & 0.5429 & 0.5479 & 0.5825 & 0.5647 \\
\midrule

\multirow{5}{*}{Pipeline B}
& None & 541 & 0.7318 & 0.6303 & 0.6773 & 0.7440 & 0.6285 & 0.6814 \\
& beoordeling & 389 & 0.5723 & 0.7326 & 0.6426 & 0.5962 & 0.7326 & 0.6574 \\
& beslissing & 69 & 0.9062 & 0.8406 & 0.8722 & 0.9062 & 0.8406 & 0.8722 \\
& materi\"ele feiten & 297 & 0.6022 & 0.5556 & 0.5779 & 0.6538 & 0.5724 & 0.6104 \\
& proceshandelingen & 206 & 0.5400 & 0.5243 & 0.5320 & 0.5391 & 0.6359 & 0.5835 \\
\midrule

\multirow{2}{*}{Stage 1}
& not-labelled / None & 541 & 0.7494 & 0.6081 & 0.6714 & 0.7494 & 0.6081 & 0.6714 \\
& labelled & 961 & 0.8006 & 0.8855 & 0.8409 & 0.8006 & 0.8855 & 0.8409 \\
\midrule

\multirow{4}{*}{Stage 2 4-way}
& beoordeling & 389 & 0.7364 & 0.8329 & 0.7817 & 0.7438 & 0.8509 & 0.7938 \\
& beslissing & 69 & 0.9677 & 0.8696 & 0.9160 & 0.9836 & 0.8696 & 0.9231 \\
& materi\"ele feiten & 297 & 0.7720 & 0.6498 & 0.7057 & 0.7976 & 0.6768 & 0.7322 \\
& proceshandelingen & 206 & 0.6029 & 0.6117 & 0.6072 & 0.6305 & 0.6214 & 0.6259 \\
\midrule

\multirow{4}{*}{OvR}
& beoordeling & 389 & 0.7616 & 0.8046 & 0.7825 & 0.8175 & 0.7943 & 0.8057 \\
& beslissing & 69 & 0.9688 & 0.8986 & 0.9323 & 0.9688 & 0.8986 & 0.9323 \\
& materi\"ele feiten & 297 & 0.7722 & 0.6734 & 0.7194 & 0.8918 & 0.5825 & 0.7047 \\
& proceshandelingen & 206 & 0.6790 & 0.5340 & 0.5978 & 0.6875 & 0.5340 & 0.6011 \\

\bottomrule
\end{tabular}%
}
\end{table*}

Given the overall model comparison, the class-level analysis focuses on RobBERT, which achieves the strongest results in all three end-to-end settings (five-way, Pipeline A, and Pipeline B), both before and after header-prior fusion, and remains close to BERTje in the component settings. Table~\ref{tab:classwise_before_after} reports class-level precision, recall, and F1 for all six RobBERT formulations, before and after header-prior fusion.

\textit{Beslissing} is the most reliably detected role when None sentences are excluded (F1 above 0.91 in both Stage 2 and OvR, with or without fusion), but markedly less reliable when these are included: F1 falls to 0.797--0.812 for the five-way model and Pipeline A, and only Pipeline B retains a stronger 0.872 — because it draws on the same independently trained detector rather than a shared multi-way classifier. This gap supports separating None Detection from role assignment: \textit{beslissing} is close to a solved role-classification problem, but remains only moderately reliable end-to-end. \textit{Beoordeling}, \textit{materi\"{e}le feiten}, and \textit{proceshandelingen} show a more consistent, modest gradient of improvement under fusion. In Stage 2, all four legal roles improve: \textit{materi\"{e}le feiten} rises from 0.706 to 0.732 F1 and \textit{proceshandelingen} from 0.607 to 0.626 (the two largest gains), \textit{beoordeling} from 0.782 to 0.794, and \textit{beslissing} from 0.916 to 0.923.

The OvR formulation shows a clear precision--recall trade-off under fusion in \textit{materi\"{e}le feiten} (precision 0.772 $\rightarrow$ 0.892, recall 0.673 $\rightarrow$ 0.583), while \textit{beoordeling} improves in both precision and F1 (0.762 $\rightarrow$ 0.818; 0.783 $\rightarrow$ 0.806). \textit{Proceshandelingen} and \textit{beslissing} are essentially unchanged, since the validation sweep found little further use for header information beyond what the text model already captures for these two detectors.


Every pipeline error can be attributed to exactly one source: the relevance stage, the role-assignment stage, or the combination formula. The direct five-way RobBERT model misclassifies 502 of 1,502 test sentences (33.4\%); 60.6\% of these are relevance errors (a relevant sentence lost to None, or vice versa) and 39.4\% confuse one legal role for another, with nearly identical proportions for Pipeline A (58.5\%/41.5\%) and Pipeline B (59.6\%/40.4\%). Among the 961 gold-relevant test sentences, about two-thirds are classified correctly by both pipelines; most remaining errors come from the role-assignment stage, fewer from Stage 1 labelling a relevant sentence as None, and under 1\% from the combination formula itself.

\section{Discussion}

The fact that the rhetorical roles in the corpus that was analysed often changed sentence-by-sentence made the classification task challenging: one must separate legally meaningful sentences from unlabelled material \textit{and} distinguish between legal roles that are not always cleanly separated. To address this, several approaches were tested: a direct five-way classifier, two staged pipelines that separate None Detection from role assignment, Bayesian header-prior fusion to inject section-header structure into predictions, and, as an alternative to fusion, a confidence-thresholding decision rule.

So which formulation performed best? The answer is: this depends on what the downstream system is meant to do with the output. Some uses are \textbf{coverage-oriented}, favouring high recall. Some noise in a summary can be acceptable, as opposed to omitting a key fact. And it presumably takes annotators less effort to remove pre-annotations than to add them. Other uses are \textbf{precision-oriented}, where noise is detrimental to the results. Retrieving statutory provisions from a case's facts is one such case: a leaked sentence of judicial reasoning can hint at the provision being searched for. Table~\ref{tab:downstream_tasks} provides examples of some representative downstream uses and matches each to the approaches displayed in Tables~\ref{app:full_supervised_results} and \ref{tab:classwise_before_after}. The table reflects our own judgement; we did not test on the downstream tasks themselves.

One limitation applies to every configuration in the table: None Detection is the harder half of the task, not role assignment. Neither header-prior fusion nor confidence thresholding improves it, and it remains the largest single source of error for the five-way model and both pipelines alike. Three more things are worth noting about the table. First, some recommendations name a single component, like Stage 2 or one OvR detector, instead of a full pipeline; using a component this way only works if something else has already filtered out irrelevant sentences first: Stage 1, a simple header rule, or a person. Second, how well a pipeline works has little to do with how its two stages are combined: the combination step causes well under 1\% of a pipeline's errors. So, which components are used matters. Third, none of the formulations tested here handle proceshandelingen well: the best F1 any of them reaches for this class is 0.63 (Table~\ref{tab:classwise_before_after}), so a task that depends specifically on identifying this class should not rely on any of these models alone; human review is still needed.


\begin{table}[!t]
\centering
\caption{Suggested approach per downstream use, based on the precision/recall profiles in Tables~\ref{app:full_supervised_results} and~\ref{tab:classwise_before_after}. The mappings reflect our judgement based on intrinsic evaluation; they were not tested on the downstream tasks themselves.}
\label{tab:downstream_tasks}
\scriptsize
\setlength{\tabcolsep}{3pt}
\renewcommand{\arraystretch}{1.15}
\begin{tabular}{l L{0.17\linewidth} L{0.16\linewidth} L{0.32\linewidth} L{0.23\linewidth}}
\toprule
\textbf{\#} & \textbf{Task} & \textbf{Objective} & \textbf{Best fits} & \textbf{Unsuitable} \\
\midrule
1 & Extractive summarization of facts & Recall on \textit{materi\"{e}le feiten} & Stage 2 alone (0.677 R) if pre-filtered; Pipeline B (0.572 R) end-to-end & Fused/precision-tuned OvR config (0.892 P / 0.583 R after pre-filtering) \\
2 & Provision retrieval from facts & Precision on \textit{feiten} & OvR \textit{materi\"{e}le feiten} detector alone (0.892 P after pre-filtering) & Five-way / Pipeline A (0.664 / 0.649 P) \\
3 & Case-based retrieval on facts & Balanced, recall-leaning & Stage 2 alone (0.732 F1) if pre-filtered; Pipeline B (0.610 F1) end-to-end & Precision-tuned OvR config \\
4 & Outcome prediction & Precision on \textit{beslissing} & Stage 2/OvR alone (0.984/0.969 P); Pipeline B (0.906 P) end-to-end & Pipeline A (0.844 P) \\
5 & Argument mining & Recall on \textit{beoordeling} & Stage 2 alone (0.851 R); Pipeline A (0.756 R) end-to-end & Precision-tuned configs \\
6 & Reading aid (highlighting roles in the text) & Balanced, no class favoured & Pipeline B, or fused five-way (0.681 / 0.677 Macro F1) & --- \\
7 & Pre-annotation for annotators & Over-select: high recall, all classes & Stage 1 (0.886 R) and Stage 2 alone, as independent signals & Precision-tuned OvR config \\
8 & RAG chunking for legal QA & Both precision and recall & Pipeline B (single-label); OvR in parallel (soft multi-label) & Pipeline A \\
\bottomrule
\end{tabular}
\end{table}

The pipeline formulation was intended to improve the classification of the rhetorical role categories. However, Pipeline B, the strongest single end-to-end predictor in this study, only modestly exceeds the direct five-way model (0.6810 vs.\ 0.6772 macro F1), not enough to justify the added complexity of a staged system. Instead, its value lies with keeping None Detection and role assignment separate and making the steps separately evaluable, which is what Table~\ref{tab:downstream_tasks} depends on. A system builder can read off precision and recall per class and per stage, then configure or replace whichever component does not fit their task. The direct five-way model cannot offer this: it is limited to one blended trade-off. Hence, the pipeline's real value is exposing the precision/recall trade-offs rather than providing a higher aggregate score. The choice of formulation is a design decision that should be made with the task in mind, and not by searching for a universally best model.

\FloatBarrier
\section{Conclusion and Future Work}

This study shows that no single classification formulation is best for sentence-level rhetorical-role classification in Dutch judicial decisions: which one to use depends on the downstream task's tolerance for missed content versus noise (Table~\ref{tab:downstream_tasks}). None Detection, not role assignment, remains the harder half of the problem. Neither header-prior fusion nor confidence thresholding improves it, but separating the two stages, as in the pipeline formulations, lets each be evaluated and deployed on its own precision/recall profile. This work contributes a Dutch sentence-level annotated dataset, a comparison of classification strategies, an error decomposition that attributes every pipeline error to exactly one component stage, and a mapping of these results onto representative downstream tasks.

Our findings are limited by a corpus of 100 decisions from three legal domains in a single jurisdiction, by intrinsic evaluation only, and by task mappings that reflect our own judgement rather than a tested deployment. Our models also classify each sentence independently; given how often roles shift between neighbouring sentences, context-aware architectures are a natural next step. Future work should also test generalisation to larger corpora, evaluate the formulations on real downstream tasks, and explore richer ways of injecting structural information than post-hoc probability fusion.
\bibliography{references}

\end{document}